\documentclass[acmtog]{acmart}
\acmSubmissionID{1422}

\usepackage[ruled]{algorithm2e} 

\SetAlFnt{\small}
\SetAlCapFnt{\small}
\SetAlCapNameFnt{\small}
\SetAlCapHSkip{0pt}

\usepackage{amsmath}
\usepackage{float}

\copyrightyear{2025}
\acmYear{2025}
\setcopyright{cc}
\setcctype{by-nc-nd}
\acmConference[SA Conference Papers '25]{SIGGRAPH Asia 2025 Conference Papers}{December 15--18, 2025}{Hong Kong, Hong Kong}
\acmBooktitle{SIGGRAPH Asia 2025 Conference Papers (SA Conference Papers '25), December 15--18, 2025, Hong Kong, Hong Kong}\acmDOI{10.1145/3757377.3763873}
\acmISBN{979-8-4007-2137-3/2025/12}

\begin{document}

\title{Learning to Refocus with Video Diffusion Models}

\author{SaiKiran Tedla}
\orcid{0000-0002-2679-2881}
\email{tedlasai@yorku.ca}
\affiliation{%
	\institution{Adobe, USA \& York University}
	\country{Canada}
}
\author{Zhoutong Zhang}
\orcid{0000-0002-8144-1611}
\email{zhoutongz@adobe.com}
\affiliation{%
	\institution{Adobe}
	\country{USA}
}
\author{Xuaner Zhang}
\orcid{0000-0002-7679-800X}
\email{cecilia77@berkeley.edu}
\affiliation{%
	\institution{Adobe}
	\country{USA}
}
\author{Shumian Xin}
\orcid{0009-0008-3974-2876}
\email{sxin@adobe.com}
\affiliation{%
	\institution{Adobe}
	\country{USA}
}

\begin{abstract}

Focus is a cornerstone of photography, yet autofocus systems often fail to capture the intended subject, and users frequently wish to adjust focus after capture. We introduce a novel method for realistic post-capture refocusing using video diffusion models. From a single defocused image, our approach generates a perceptually accurate focal stack, represented as a video sequence, enabling interactive refocusing and unlocking a range of downstream applications. We release a large-scale focal stack dataset acquired under diverse real-world smartphone conditions to support this work and future research. Our method consistently outperforms existing approaches in both perceptual quality and robustness across challenging scenarios, paving the way for more advanced focus-editing capabilities in everyday photography. Code and data are available at \href{https://learn2refocus.github.io/}{\texttt{www.learn2refocus.github.io}}

\end{abstract}

\ccsdesc[500]{Computing methodologies~Computational photography}
\ccsdesc[500]{Computing methodologies~Image processing}

\keywords{Image refocusing; video diffusion models}
 
\begin{teaserfigure}
  \centering
  \includegraphics[width=\linewidth]{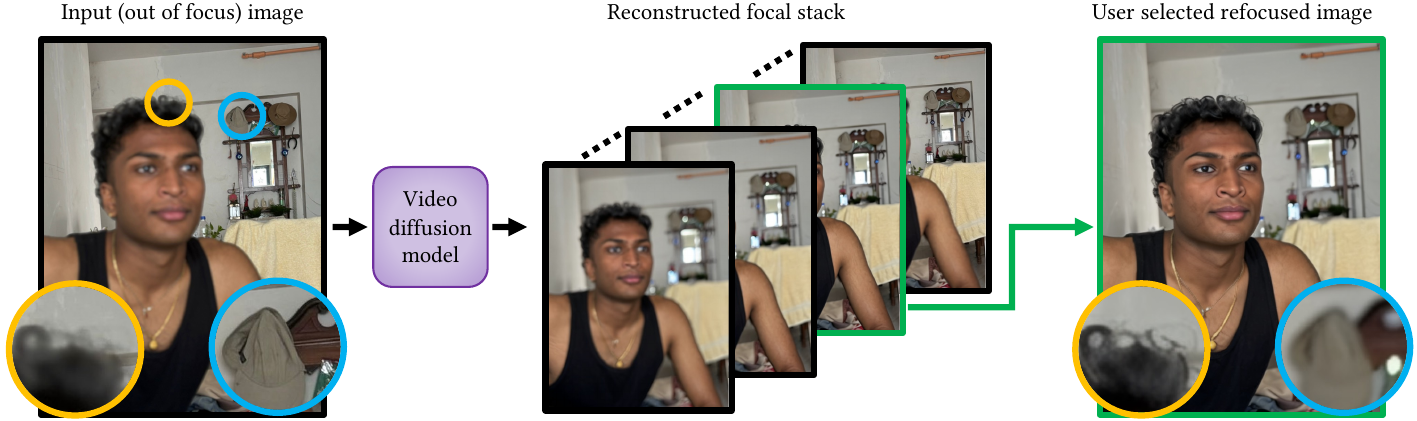}
  \caption{We present a method for refocusing images using video diffusion models to predict focal stacks, allowing users to select their preferred focus. Trained on a large smartphone dataset captured in diverse real-world environments, our model generates realistic defocus effects and performs effectively on real scenes.}
  \label{fig:teaser}
\end{teaserfigure}

\maketitle


\newcommand{\projectpath}{https://github.com/CompVis/latent-diffusion}

\newcommand*{\approxident}{%
  \mathrel{\vcenter{\offinterlineskip
  \hbox{$\sim$}\vskip-.35ex\hbox{$\sim$}\vskip-.35ex\hbox{$\sim$}}}}
\newcommand{\RR}{\mathbb{R}}
\newcommand{\expec}{\mathbb{E}}
\newcommand{\prior}{\mathcal{N}(0,1)}

\newcommand{\xpixel}{x_\text{pixel}}
\newcommand{\xpixelrec}{\tilde{x}_\text{pixel}}
\newcommand{\hpixel}{H}
\newcommand{\wpixel}{W}
\newcommand{\cpixel}{3}
\newcommand{\xrec}{\tilde{x}}

\newcommand{\latent}{z_0}
\newcommand{\hlatent}{h}
\newcommand{\wlatent}{w}
\newcommand{\clatent}{c}

\newcommand{\zt}[1]{z_{#1}}
\newcommand{\enc}{E}
\newcommand{\ppixel}{p_\text{pixel}}
\newcommand{\pzt}[1]{p_{\zt{#1}}}
\newcommand{\qzt}[1]{q_{\zt{#1}}}
\newcommand{\q}{q}
\newcommand{\R}{\mathbb{R}}

\newcommand{\LPIPS}{\text{LPIPS}}
\newcommand{\KL}{\mathbb{KL}}
\newcommand{\expect}{\mathbb{E}}
\newcommand{\pmodel}[1]{p^{#1}_{\theta}}
\newcommand{\pchain}{p_{\theta}}
\newcommand{\qchain}{q}
\newcommand{\qmodel}[1]{q_{#1}}

\newcommand{\qenc}{q_{\phi}}
\newcommand{\pdec}{p_{\phi}}
\newcommand{\dec}{G_{\phi}}

\newcommand{\disc}{D_{\psi}}

\newcommand{\lrec}{L_{rec}}
\newcommand{\ladv}{L_{adv}}
\newcommand{\lreg}{L_{reg}}
\newcommand{\lcomp}{L_{cm}}
\newcommand{\lsimple}{L_{DM}}
\newcommand{\lsimpleldm}{L_{LDM}}
\newcommand{\lsimplelcm}{L_{LDM}}

\newcommand{\model}{\epsilon_\theta}
\newcommand{\conditioner}{\tau_\theta}

\newcommand{\encoder}{\mathcal{E}}
\newcommand{\decoder}{\mathcal{D}}

\newcommand{\cond}{y}

\section{Introduction}
Focus is a pivotal element of photography, yet modern autofocus systems can easily fail in dynamic or complex scenes. Photographers frequently wish to shift the focal plane post-capture, whether for aesthetic or corrective reasons. Existing techniques for adjusting focus typically require specialized hardware such as light field cameras~\cite{ng2005light, zhang2018lightfield, levoy2006light} or rely on multiple images~\cite{hadi, abuolaim2020defocus} and depth maps~\cite{fiss2014refocusing, moreno2007active}. While these methods are effective, they demand resources and expertise beyond the reach of most users. Moreover, many approaches focus on producing all-in-focus images~\cite{restormer, naf, ifan}, which do not replicate the natural defocus effects found in optical systems.

In this paper, we present a novel method for post-capture refocusing that requires only a single defocused image and leverages video diffusion models. As illustrated in Figure~\ref{fig:teaser}, our approach produces a realistic focal stack---a sequence of images with progressively shifting focus---enabling interactive refocusing via a simple slider. Refocusing, especially from severe blur, is inherently ill-posed. This observation prompts us to treat it as a generative task. We cast refocusing as a multi-frame generation problem, where the focal stack is modeled as a temporally coherent video sequence.

\begin{figure}[t]
  \centering
  \includegraphics[width=\linewidth]{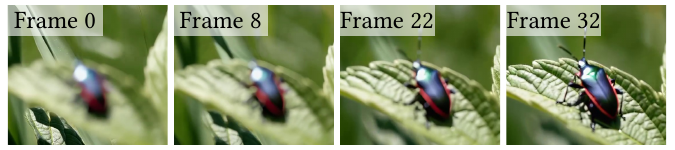}
  \caption{Illustration showing that video diffusion models have priors on generating focal stacks. The video sequence is generated by prompting a model~\cite{firefly} to generate ``a focus pull of a bug on a leaf''.}
  \label{fig:videodiffusionprior}
\end{figure}

As seen in Figure~\ref{fig:videodiffusionprior}, we observe that pre-trained video diffusion models have priors for doing focus pulls. We hypothesize these priors are helpful for performing refocusing in a variety of scenarios. Thus, we propose to use video diffusion models to reconstruct the entire focal stack conditioned on just a single defocused input. By introducing a minimal modification to the classifier-free guidance mechanism, our method consistently yields perceptually realistic focal stacks across a range of challenging scenarios.

To facilitate the development and evaluation of our approach, we introduce a new large-scale focal stack dataset containing 1,637 samples captured on an iPhone 12 under a variety of real-world conditions. This dataset offers a practical benchmark for refocusing techniques, covering diverse scenes and lighting scenarios. Although its size is relatively modest, we leverage the strong priors embedded in pre-trained video diffusion models to achieve robust performance.

Our method consistently produces perceptually realistic results and remains resilient across challenging capture conditions. By bridging state-of-the-art generative modeling with everyday smartphone photography, we deliver advanced refocusing capabilities to a broad user base, expanding the potential for post-capture editing.

We summarize our contributions as follows:
\begin{itemize}
    \item In Section ~\ref{sec:dataset}, we introduce a large-scale focal stack dataset of real-world scenes captured on a modern smartphone.
    \item In Section ~\ref{sec:method}, we propose a novel approach to adapt video diffusion models for refocusing by casting the task as focal stack prediction.
    \item In Section ~\ref{sec:results}, we demonstrate that our method allows realistic refocusing, enabling both defocus deblurring and natural defocus effects.
\end{itemize}

\section{Related work} \label{sec:related_work}
In this section, we review existing methods for manipulating focus and depth of field in images and then discuss why diffusion models provide a compelling solution for refocusing.

\subsection{Focus and depth-of-field manipulation}

\paragraph{Image refocusing.} The primary prior work addressing refocusing from a single image is RefocusGAN~\cite{refocusgan}. It employs a two-stage approach: first, predicting an all-in-focus (AiF) image, then generating a defocused image at the desired focus. RefocusGAN can construct a full focal stack by iteratively applying the second stage. RefocusGAN is trained exclusively on synthetic data.
In contrast, our method directly produces full focal stacks in one pass and is trained on real-world smartphone captures, leading to greater realism in everyday photography scenarios. 

\paragraph{Light field photography.} 
Light field imaging captures the spatial-directional information of light rays in a scene, enabling post-capture refocusing~\cite{ng2005light, ruan2021aifnet, levoy2006light}. However, such methods require specialized hardware and are less practical for mainstream consumer photography.

\paragraph{Defocus deblurring and defocus blur generation.} 
Defocus deblurring techniques aim to transform out-of-focus images into AiF versions using standard~\cite{naf, restormer, ifan, wang2022uformer, ruan2024self} or diffusion-based restoration networks~\cite{chen2025efficient, kong2025deblurdiff, feng2025residual}. Other methods utilize dual-pixel data and multiplane representations to enable refocusing~\cite{xin}. On the opposite end, bokeh and defocus generation approaches rely on an AiF image and a depth map to produce defocused results~\cite{bokeh1, bokeh2, peng2022bokehme}. Alzayer \textit{et al.}~\cite{hadi} instead use dual captures (wide and ultrawide) to jointly estimate an AiF image, depth map, and editing control. In contrast, our goal is to generate a full focal stack from a single defocused image, eliminating the need for multiple inputs or sensor-specific data.

\paragraph{Controllable image generation.} Recently work~\cite{fang2024camera, generative_photography} has explored controllable text-to-image generation conditioned on camera parameters, such as focal length, aperture, and shutter speed. Fang \textit{et al.}~\cite{fang2024camera} briefly discuss refocusing via SDEdit~\cite{sdedit}, the method does not consistently preserve content identity and often alters the original scene. 


\paragraph{Focal stack and light field datasets.} Most prior methods for depth-of-field manipulation~\cite{peng2022bokehme, sheng2024dr, levoy2006light, refocusgan} use synthetic focal stacks or light-field data, which may not reflect real-world photography conditions. Currently, the largest publicly available real-world dataset~\cite{learningtoautofocus} contains 510 focal stacks captured with a Google Pixel 3. To advance research in this domain, we introduce a new large-scale dataset comprising 1637 focal stacks captured under diverse real-world scenarios using an iPhone 12.

\subsection{Diffusion models}
\paragraph{Generative image enhancement and editing.} Diffusion models have demonstrated their effectiveness in tasks such as image restoration and generation, including super-resolution~\cite{fleetsuperresolution} and impainting~\cite{deblur_and_impaint, impaintingdiffusion}. Similar to our refocusing task, diffusion models can be used for deblurring~\cite{deblurring1,deblur_and_impaint, chen2025efficient}; however, these methods typically only convert a blurred image into an all-in-focus image. Thus, they require modifications or additional conditioning mechanisms to predict focal stacks.
Diffusion models are particularly powerful for image editing, leveraging text-based controls~\cite{edit1, edit2} or parameter sliders~\cite{guerrero2024texsliders}. Their generative flexibility makes them especially appealing for creative edits or tasks involving significant content restoration.


\paragraph{Generative camera control.} Our work is inspired by those that fine-tune video diffusion models for pose-conditioned generation~\cite{cameractrl1, vd3d}. These works contain no explicit 3D constraints and only rely on implicit priors within the video model. Our approach also models focal stacks implicitly by utilizing the prior from video diffusion models and our data.

\begin{figure*}
  \centering
  \includegraphics[width=1.0\linewidth]{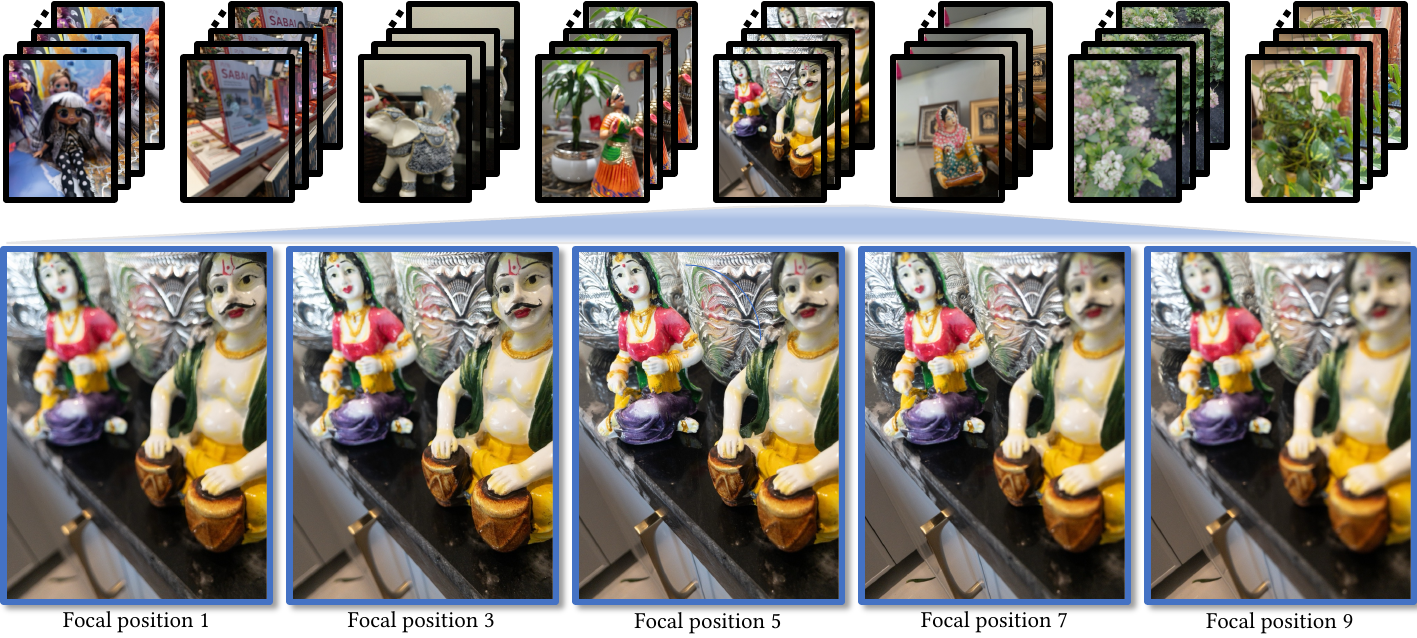}
  \caption{Eight sample focal stacks from our dataset consisting of 1637 total scenes. More focal stacks are visualized in a supplementary material montage.} 
  \label{fig:scenes}
\end{figure*}

\paragraph{Generative video interpolation.} Our approach treats refocusing as a video-generation problem, partly inspired by generative video interpolation research. For instance, pixel-based video diffusion models can interpolate between given start and end frames~\cite{vidim}, while latent video diffusion methods incorporate autoencoder features~\cite{ldmvfi} or motion priors~\cite{huang2024motion}. These methods primarily focus on frame interpolation, whereas our task requires extrapolation from a single defocused image to produce an entire focal stack.  


\section{Large focal stack dataset}
\label{sec:dataset}

We introduce a new focal stack dataset that, to our knowledge, is the largest publicly available collection of real-world smartphone focal stacks.

\subsection{Data collection}
\begin{figure}[h!]
  \centering
  \includegraphics[width=0.55\linewidth]{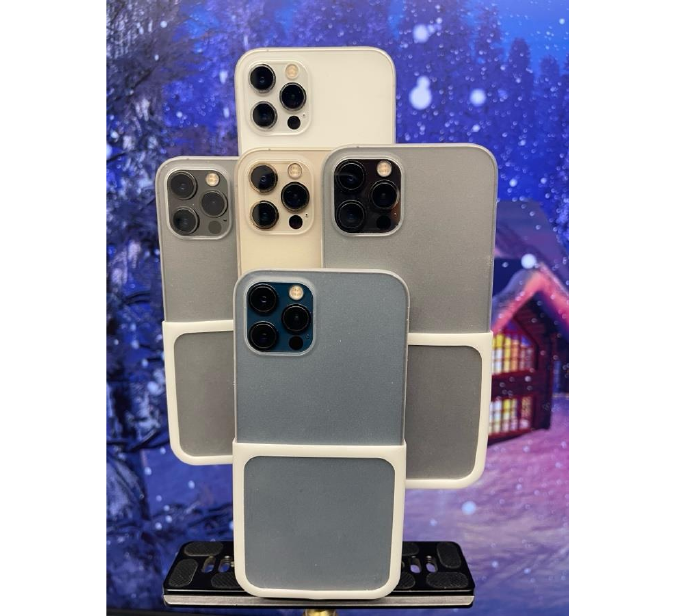}
  \caption{The rig used to capture our dataset. For this work, we only use the central camera focal stacks.}
  \label{fig:rig}
\end{figure}

The dataset consists of 1,637 scenes, each captured with a custom rig of five iPhones (Figure~\ref{fig:rig}) similar to Herrmann et al.~\cite{learningtoautofocus}. Each phone simultaneously records a focal stack and an ultrawide shot for each scene, creating a rich source of multi-view data. We collect focal stacks by moving the iPhone’s focus from its minimum to maximum distance in nine increments. In the iPhone API, this corresponds to focus values from 0 to 0.8, in steps of 0.1. Although the API supports values up to 1.0, we found 0.8 to be the typical upper bound in practical scenarios. For the current work, we leverage only the central camera’s focal stack. However, the full dataset (including ultrawide captures and all cameras) will be publicly available to support future research needing depth or stereo information.

The dataset comprises 1,637 scenes split into 1,474 training and 163 test focal stacks. A large portion focuses on macro photography, where the shallow depth-of-field enables the richest focal shifts on the iPhone. To ensure diversity and realism, we capture both controlled and in-the-wild scenes. Controlled shots are taken in lightboxes with fixed cameras, stable lighting, and minimal scene motion, while natural scenes come from varied environments (e.g., homes, bookstores, universities) with changing lighting and slight background motion.
All images are captured in RAW format at a resolution of $4032 \times 3024$, maintaining consistent exposure and white balance settings.

\begin{figure}
  \centering
  \includegraphics[width=1.0\linewidth]{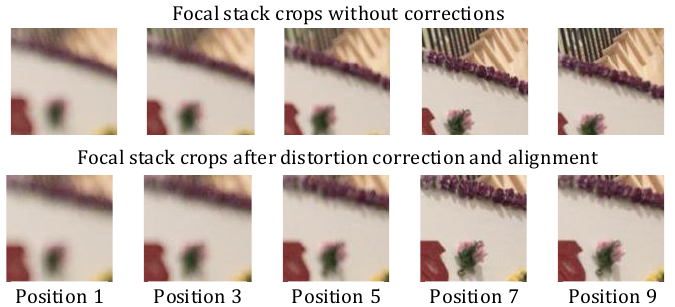}
  \caption{ The original focal stack exhibits misalignment caused by focal breathing, evident from the noticeable edge movements. Our pre-processing corrects these distortions, resulting in a well-aligned stack that enables more accurate refocusing.}
  \label{fig:breathing_correction}
\end{figure}

\subsection{Data processing}
We apply post-processing to ensure high-quality alignment and realism within each focal stack. The initial focal stack contains focal breathing, resulting in slight changes in the field of view and distortion at each focal position. We perform a careful processing to remove these focal breathing effects as seen in Figure~\ref{fig:breathing_correction}, because without this correction, the model would have to learn how to perform field of view changes and distortion corrections in addition to refocusing. In the supplementary, we show that our model and baselines fail to attain good refocusing results without our processing steps.

\paragraph{Raw conversion.} We begin by converting each RAW image to the sRGB color space using Adobe Camera Raw (ACR)~\cite{acr}.

\paragraph{Distortion correction.} Next, we apply a calibrated radial distortion profile per focus setting to correct lens aberrations at each step of the focal stack.

\paragraph{Alignment.} Due to focal breathing, images are not perfectly aligned after distortion correction. We address this by applying a pre-calculated scale transformation so that all images in the stack align with the first frame (which has the smallest field of view). 

\paragraph{All-in-focus image generation.}
Finally, we generate an AiF image for each stack using Helicon Focus~\cite{heliconfocus} in depth mode, providing a high-quality reference for evaluating defocus deblurring methods.

After these processing steps, each scene in our dataset is represented by nine aligned focal stack images and one AiF image, offering a comprehensive benchmark for real-world refocusing research. A sampling of scenes is shown in Figure~\ref{fig:scenes}, and more scenes are visualized in the supplementary material dataset montage.


\section{Method}
\label{sec:method}
We aim to synthesize a perceptually realistic focal stack from a single defocused image. Key to our approach is the observation that a focal stack can be viewed as a video, where each frame corresponds to a different focus distance. Leveraging this perspective, we employ a video diffusion model to jointly reconstruct the entire focal stack in one sampling process. This contrasts RefocusGAN~\cite{refocusgan}, which refocuses one frame at a time.

In Section~\ref{sec:method_preliminaries}, we provide an overview of diffusion models as the foundation of our work. In Section~\ref{sec:method_modified_cfg}, we detail our modifications to the classifier-free guidance (CFG) mechanism of a state-of-the-art video diffusion model~\cite{svd}, enabling robust and realistic single-shot generation of focal stacks.

\subsection{Preliminaries} \label{sec:method_preliminaries}

\paragraph{Latent diffusion models.}
Diffusion models learn data distributions by reversing a forward Gaussian noising process. While pixel diffusion models~\cite{ho2022video} operate directly in the high-dimensional pixel space, latent diffusion models~\cite{svd} work within a compressed latent space~$\boldsymbol{z}$, generated by a pre-trained variational auto-encoder (VAE)~\cite{sd}. This significantly reduces computational overhead, especially for processing video data. In this work, we adopt Stable Video Diffusion (SVD)~\cite{svd} for its simplicity, effectiveness, and public availability. 

In latent diffusion models (LDMs), the denoiser~$\model$ is trained to reverse the noising process at each timestep~$t$. Let~$\boldsymbol{z}_t$ be the noisy latent state at timestep~$t$, and let $\epsilon \sim \mathcal{N}(0, 1)$ be the Gaussian noise. The training objective is to minimize the discrepancy between the predicted noise and the true noise:
\begin{equation}
	\lsimpleldm := \expec_{\epsilon \sim \mathcal{N}(0, 1), \boldsymbol{z}_t, t}\Big[ \Vert \epsilon - \model(\boldsymbol{z}_t,t) \Vert_{2}^{2}\Big] \, .
	\label{eq:ldmloss}
\end{equation}

 
\paragraph{Classifier-free guidance.}
Classifier-free guidance (CFG)~\cite{cfg} is a mechanism that allows diffusion models to incorporate additional conditioning signals, such as a starting frame or a text prompt. In latent video diffusion models, the denoiser~$\model(\boldsymbol{z_{t}}, \boldsymbol{c}, t)$ is trained with a conditioning input $\boldsymbol{c}$. During sampling, the predicted noise $\widetilde{\model}(\boldsymbol{z}_{t}, \boldsymbol{c}, t) $ is computed as a linear combination of conditional and unconditional noise estimates:
\begin{equation}
  \widetilde{\model}(\boldsymbol{z}_{t}, \boldsymbol{c}, t)  = (1+w)\model(\boldsymbol{z}_{t},\boldsymbol{c}, t) - w\model(\boldsymbol{z}_{t}, t) ,
  \label{eq:cfg_ori}
  \end{equation}
where $w$ is a weight scalar that balances the influence of the conditioned and unconditioned outputs.

In SVD, the video latent $\boldsymbol{z}_t$ is a sequence of per-frame latents $\boldsymbol{z}_t = [\boldsymbol{f}_1 , \dots, \boldsymbol{f}_F]$, where $\boldsymbol{f}_p$ represents the latent for frame $p$ in a video with $F$ frames. 
The model conditions on the first frame latent $\boldsymbol{f}_1$ by replicating it across all frames:
\begin{equation}
  \widetilde{\model} 
  = (1+w)\model(\boldsymbol{z}_{t},[\boldsymbol{f}_1 , \dots, \boldsymbol{f}_1], t)  - w\model(\boldsymbol{z}_{t} ,[\boldsymbol{0}, \dots, \boldsymbol{0}], t) .
  \label{eq:cfg}
\end{equation}
This approach enables a single network to handle both conditioned and unconditioned noise estimates by zeroing out the conditioning tensor for the unconditioned case.

\subsection{Modified classifier-free guidance} \label{sec:method_modified_cfg}

Our goal is to generate the entire focal stack from a single image. To accomplish this, we condition a video diffusion model on one selected focal position from a fixed-length stack in our dataset, and then, in a single sampling pass, predict all frames including the selected one.

In the original SVD, conditioning is restricted to the first frame by replicating its latent across all frames, effectively ignoring positional information. This replication introduces ambiguity: if a focal position  $\boldsymbol{f}_p$ is chosen randomly for each training batch and duplicated across the entire stack, the model sees inconsistent focal positions and struggles to learn a clear mapping between the input and output focal positions.

\begin{figure}[t]
  \centering
  \includegraphics[width=0.5\textwidth]{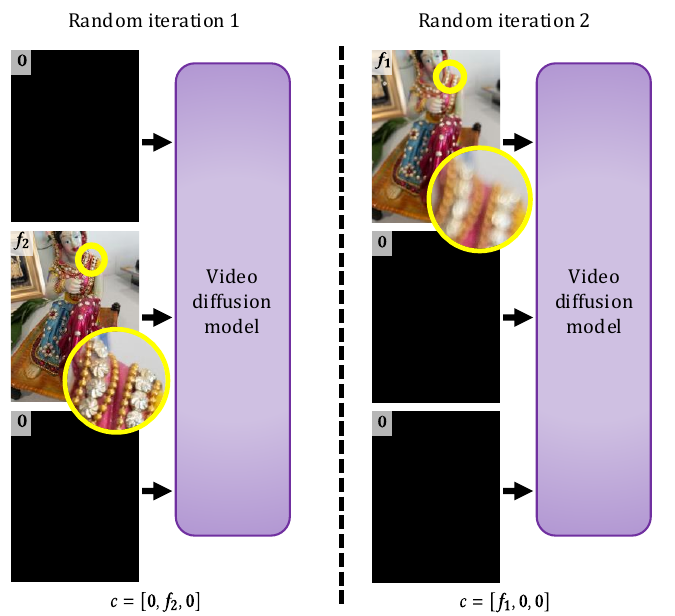}
  \caption{Illustration of our modified classifier-free guidance for a focal stack with $F=3$ frames. During training, the latent corresponding to a randomly selected focal position is passed to the video diffusion model at the matching position, while all other conditioning inputs are set to zero.}
  \label{fig:guidance}
\end{figure}

To overcome this limitation, we introduce \textit{position-dependent} conditioning.
Specifically, rather than replicating the chosen latent $\boldsymbol{f}_p$ across all positions, we place it only at the corresponding position $p$ in the conditioning tensor, while setting all other positions to zero (Figure~\ref{fig:guidance}). Formally, for focal position $p$:
 \begin{equation}
  \widetilde{\model}
  = (1+w)\model(\boldsymbol{z}_{t},[\boldsymbol{0}, \dots, \boldsymbol{f}_p, \dots, \boldsymbol{0}], t) - w\model(\boldsymbol{z}_{t} ,[\boldsymbol{0}, \dots, \boldsymbol{0}], t) .
  \label{eq:cfg_modify}
   \end{equation}

Our approach takes inspiration from VIDIM~\cite{vidim}, which concatenates frames along the time axis for video interpolation and treats conditioning frames as part of the input sequence. However, unlike interpolation methods that typically assume fixed start and end frames, our task demands flexibility, as the input frame can appear at any position in the focal stack. By ensuring the UNet~\cite{svd} learns to reconstruct focal stacks aligned with the input focal position, our modification enables position-dependent refocusing.

During training, we randomly select focal positions $p$ for each batch, ensuring the model covers the entire range of the stack. This approach effectively leverages the known focal stack structure while introducing only minor modifications to the existing diffusion framework. Architecture details are given in the supplementary.

\section{Experiments} \label{sec:results}
We evaluate our method across diverse tasks to demonstrate its perceptual realism, share-readiness, and versatility. Section~\ref{sec:experiments_implementation} details implementation and comparisons with state-of-the-art methods. Section~\ref{sec:experiments_our_dataset} presents qualitative, quantitative, and ablation results on our dataset. Section~\ref{sec:experiments_everyday_user} tests real-world use with user-provided images. Section~\ref{sec:experiments_other_cameras} examines generalization to various camera types. Section~\ref{sec:experiments_downstream} explores downstream tasks like all-in-focus generation, synthetic DoF, and motion deblurring. Section~\ref{sec:experiments_limitations} discusses limitations and future directions.


\subsection{Implementation details} \label{sec:experiments_implementation}
We compare our method against two reimplementations of RefocusGAN~\cite{refocusgan} and modified deblurring networks~\cite{naf, restormer}. Since RefocusGAN lacks publicly available code, we reimplemented its two-stage pipeline: the first network constructs the all-in-focus image, and the second network performs blurring, conditioned on the all-in-focus image, input image, and a focal position. We evaluate two versions of RefocusGAN: one using NAFNet~\cite{naf} (RGAN-N) and another using Restormer~\cite{restormer} (RGAN-R). 
We also implement a strong baseline using minimally modified (just edits to first/last convolutions) versions of these two deblurring networks tailored to our problem. Standard deblurring networks don't predict focal stacks and instead accept 3-channel inputs and predict an all-in-focus image. Thus, we modify the input to be a sparse focal stack concatenated along the channel dimension, with only one non-zero image, and the output to be a full focal stack. Finally, as image diffusion baselines, we use InstructPix2Pix (IP2P)~\cite{instructpix2pix} and Swintormer~\cite{chen2025efficient}. IP2P utilizes prompts for editing control. We train IP2P by providing input and output focus paired images along with prompts such as “Change focal position 1 to focal position 4”. Swintormer is minimally modified to allow for focal stack prediction. More details are given in the supplementary.

All networks are trained for 200,000 epochs with a batch size of 4 using our large focal stack dataset and the same splits. Each batch element uses a randomly selected focal position index for the input frame. For the RefocusGAN reimplementations, we train both networks for 200,000 epochs. Training is performed on full images resized to $896 \times 640$ pixels rather than patches to account for the slight shifts in radial distortion that occur within the stack, even with our data processing pipeline. 

We use the Adam optimizer for all methods. For the baseline methods, we use a learning rate of 0.0001 and an L2 loss. Our method uses a learning rate of 0.0004 and trains the denoising process on a random timestep per iteration, also with an L2 loss. During training, we retrain only the UNet transformer blocks in our model and and apply an exponential moving average (EMA)~\cite{svd} with a decay rate of 0.00001. At test time, we sample our model once using a CFG weighting of 1.5. For the IP2P baseline, we performed a grid search to identify the best hyperparameters, selecting a text guidance scale of 7.5 and an image guidance scale of 3.0.

\begin{figure*}[h!]
\centering
\includegraphics[width=\linewidth]{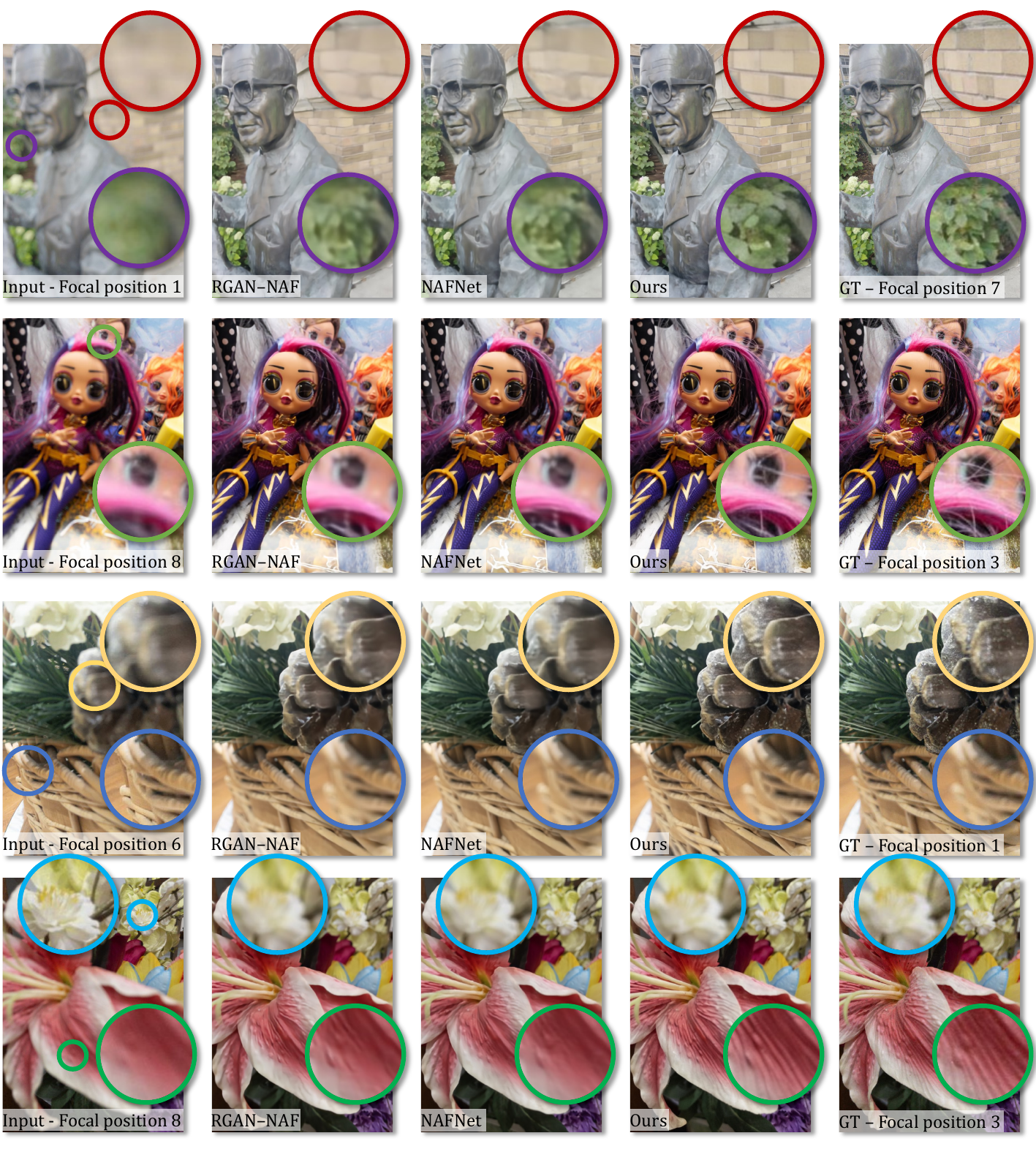}
\caption{Refocusing results from various input focal positions to output focal positions. We compare our method to RGAN-NAF~\cite{refocusgan}, NAFNet~\cite{naf}, and ground truth (GT). Our method reconstructs coarse structures in the pine-cone (third row) and fine details in hair (second row).}
\label{fig:qualitative_results}

\end{figure*}

\subsection{Results on our dataset}
\label{sec:experiments_our_dataset}

\paragraph{Qualitative results.}
We visualize sample results in Figure~\ref{fig:qualitative_results} and use NAFNet~\cite{naf} for comparison (additional comparisons with NAFNet and IP2P are available in our supplementary demo) as it outperformed other baselines in most scenarios.  The generative nature of our model enables it to generate/hallucinate high-frequency details more effectively than baseline methods.
For instance, in the second row, the doll's hairs appears more realistic and are not blurred into the background as seen in the baselines. Similarly, coarser structures, such as the bricks in the background of the first row and the pine cone in the third row, are reconstructed with better perceptual realism.
By leveraging generative modeling and score-based objectives, our method utilizes strong priors about the world to deblur and introduce realistic defocus. In contrast, baseline methods suffer from a regression-to-the-mean effect as described by Milanfar and Delbracio~\shortcite{delbracio2023inversion}. This effect occurs when refocusing over larger focal position jumps, where multiple plausible solutions exist for the refocused image. As a result, non-generative methods average these solutions, resulting in the blurring artifacts observed in the baseline results in Figure~\ref{fig:qualitative_results}.

\begin{table*}[]
  \caption{Quantitative evaluations of refocusing methods on our focal stack dataset. Best results are highlighted in bold.}
  \footnotesize
  \begin{center}
  \begin{tabular}{lllllllllllllllllllllllll}
  \toprule
  & \multicolumn{18}{c}{Refocusing from focal position 1\hspace{0 mm} }\\
                        & \multicolumn{9}{c}{LPIPS $\downarrow$}   & \multicolumn{9}{c}{FID $\downarrow$}     \\
                       
                        \cmidrule{1-19}
  Method & \multicolumn{1}{l}{F1} &\multicolumn{1}{l}{F2} &\multicolumn{1}{l}{F3} &\multicolumn{1}{l}{F4} & \multicolumn{1}{l}{F5} & \multicolumn{1}{l}{F6} & \multicolumn{1}{l}{F7} &\multicolumn{1}{l}{F8} & \multicolumn{1}{l|}{F9}& \multicolumn{1}{l}{F1} &\multicolumn{1}{l}{F2} &\multicolumn{1}{l}{F3} &\multicolumn{1}{l}{F4} & \multicolumn{1}{l}{F5} & \multicolumn{1}{l}{F6} & \multicolumn{1}{l}{F7} &\multicolumn{1}{l}{F8} & \multicolumn{1}{l}{F9}  \\ 
  \midrule
  Ablation & 0.246 & 0.359 & 0.409 & 0.452 & 0.485 & 0.508 & 0.514 & 0.504 & 0.488 & 59.40 & 87.13 & 102.5 & 114.7 & 125.1 & 133.3 & 137.5 & 134.0 & 126.7 \\
 
  IP2P & 0.163 & 0.194 & 0.224 & 0.262 & 0.289 & 0.306 & 0.305 & 0.288 & 0.262 & 26.38 & 26.25 & 32.14 & 40.88 & 47.19 & 50.60  & 52.59 & 51.31 & 46.19 \\
  
  Swintormer           & \textbf{0.006} & 0.161 & 0.190 & 0.218 & 0.249 & 0.277 & 0.285 & 0.268 & 0.239 & 0.540 & 26.19 & 34.12 & 40.68 & 48.51 & 56.07 & 59.23 & 55.96 & 49.38 \\
  RGAN-Rest & 0.008 & 0.157 & 0.186 & 0.214 & 0.245 & 0.274 & 0.282 & 0.265 & 0.235 & 0.61  & 24.74 & 32.94 & 39.45 & 47.44 & 55.80 & 60.76 & 58.32 & 51.57 \\
  RGAN-NAF    & 0.009 & 0.149 & 0.179 & 0.203 & 0.232 & 0.259 & 0.266 & 0.250 & 0.223 & 0.60  & 22.43 & 32.05 & 37.78 & 45.49 & 53.36 & 57.83 & 56.21 & 50.56 \\
  Restormer            & \textbf{0.006} & 0.144 & 0.173 & 0.201 & 0.232 & 0.260 & 0.268 & 0.249 & 0.220 & 0.57  & 20.69 & 30.68 & 37.76 & 45.78 & 53.17 & 57.26 & 54.87 & 47.51 \\
  NAFNet               & \textbf{0.006} & \textbf{0.140} & 0.170 & 0.195 & 0.224 & 0.251 & 0.259 & 0.242 & 0.213 & \textbf{0.52}  & \textbf{20.44} & 31.42 & 37.65 & 44.44 & 51.51 & 55.42 & 53.44 & 46.33 \\

  Ours                 & 0.088 & 0.141 & \textbf{0.165} & \textbf{0.190} & \textbf{0.212} & \textbf{0.229} & \textbf{0.232} & \textbf{0.219} & \textbf{0.196} & 10.76 & 21.20 & \textbf{27.40} & \textbf{31.78} & \textbf{37.16} & \textbf{42.61} & \textbf{44.67} & \textbf{43.39} & \textbf{36.34} \\
  \midrule
  & \multicolumn{18}{c}{Refocusing from focal position 5 \hspace{0 mm} }\\
  \midrule
 Ablation & 0.363 & 0.313 & 0.260 & 0.210 & 0.174 & 0.232 & 0.275 & 0.298 & 0.315 & 81.30 & 66.99 & 53.02 & 38.21 & 32.58 & 43.89 & 55.85 & 64.35 & 74.36 \\

  IP2P & 0.219 & 0.217 & 0.216 & 0.214 & 0.214 & 0.238 & 0.253 & 0.250 & 0.232 & 37.85 & 35.54 & 34.20  & 30.48 & 30.78 & 32.81 & 38.87 & 42.82 & 39.58 \\
  Swintormer           & 0.188 & 0.169 & 0.151 & 0.125 & \textbf{0.003} & 0.134 & 0.169 & 0.176 & 0.170 & 0.260 & 37.59 & 30.42 & 23.02 & 14.72 & 15.58 & 23.80 & 27.64 & 28.18 \\

  RGAN-Rest & 0.175 & 0.160 & 0.146 & 0.120 & \textbf{0.003} & 0.132 & 0.169 & 0.177 & 0.170 & 33.26 & 28.56 & 22.72 & 14.05 & \textbf{0.19}  & 14.99 & 24.19 & 29.06 & 30.20 \\
  RGAN-NAF    & 0.169 & 0.155 & 0.142 & 0.118 & 0.005 & 0.126 & 0.160 & 0.166 & 0.160 & 31.19 & 27.85 & 21.77 & 13.46 & 0.25  & 13.61 & 22.29 & 26.89 & 28.02 \\
  Restormer            & 0.164 & 0.147 & 0.134 & 0.111 & 0.006 & 0.124 & 0.160 & 0.164 & 0.157 & 28.37 & 24.40 & 19.81 & 11.44 & 0.61  & 12.08 & 21.77 & 25.70 & 25.59 \\
  NAFNet               & 0.159 & 0.144 & \textbf{0.131} & \textbf{0.106} & 0.005 & \textbf{0.118} & \textbf{0.156} & \textbf{0.160} & \textbf{0.154} & 26.86 & 23.27 & \textbf{19.07} & \textbf{10.57} & 0.43  & \textbf{11.74} & \textbf{21.75} & \textbf{25.11} & \textbf{25.17} \\

  Ours                 & \textbf{0.144} & \textbf{0.139} & 0.139 & 0.138 & 0.115 & 0.160 & 0.171 & 0.168 & 0.156 & \textbf{23.35} & \textbf{21.77} & 21.56 & 20.17 & 16.63 & 23.67 & 26.48 & 27.54 & 25.62 \\
  \midrule
  & \multicolumn{18}{c}{Refocusing from focal position 9 \hspace{0 mm} }\\
  \midrule
Ablation & 0.368 & 0.347 & 0.334 & 0.329 & 0.328 & 0.328 & 0.310 & 0.276 & 0.279 & 90.06 & 84.49 & 83.16 & 82.89 & 82.76 & 80.74 & 73.82  & 63.75 & 66.35 \\

IP2P & 0.228 & 0.229 & 0.239 & 0.251 & 0.263 & 0.269 & 0.256 & 0.226 & 0.189 & 41.16 & 39.48 & 41.01 & 42.69 & 42.57 & 41.67 & 37.82 & 32.11 & 28.73 \\
Swintormer           & 0.209 & 0.203 & 0.201 & 0.203 & 0.208 & 0.207 & 0.187 & 0.143 & \textbf{0.004} & 0.280 & 45.05 & 42.82 & 41.93 & 40.73 & 38.30 & 34.93 & 28.30 & 18.32 \\

  RGAN-R & 0.199 & 0.194 & 0.196 & 0.201 & 0.209 & 0.209 & 0.189 & 0.143 & \textbf{0.004} & 45.76 & 45.32 & 44.68 & 44.16 & 41.32 & 36.83 & 29.36 & 17.39 & 0.33  \\
RGAN-N    & 0.196 & 0.190 & 0.191 & 0.196 & 0.203 & 0.203 & 0.183 & 0.133 & 0.006 & 43.65 & 42.88 & 43.24 & 42.77 & 40.52 & 36.43 & 28.36 & \textbf{15.93} & \textbf{0.27}  \\
Restormer            & 0.188 & 0.183 & 0.184 & 0.191 & 0.200 & 0.203 & 0.186 & 0.135 & 0.006 & 40.74 & 40.36 & 40.70 & 40.62 & 38.94 & 35.89 & 29.36 & 16.37 & 0.55  \\
NAFNet               & 0.184 & 0.178 & 0.179 & 0.185 & 0.193 & 0.196 & \textbf{0.179} & \textbf{0.133} & 0.005 & 37.57 & 36.79 & 37.07 & 37.91 & 36.53 & 33.95 & \textbf{27.76} & 16.27 & 0.43  \\

Ours                 & \textbf{0.156} & \textbf{0.157} & \textbf{0.164} & \textbf{0.174} & \textbf{0.185} & \textbf{0.192} & 0.183 & 0.158 & 0.010 & \textbf{27.37} & \textbf{28.05} & \textbf{28.81} & \textbf{29.33} & \textbf{29.84} & \textbf{30.69} & 28.71 & 24.33 & 12.81 \\

  \bottomrule
  \end{tabular}
  \end{center}
  \label{tab:main_benchmark}
  \end{table*}

\paragraph{Quantitative results.}
Our evaluation focuses on the perceptual quality of refocused images, primarily using LPIPS and FID metrics. We utilize FID as it is a common metric in generative tasks that measures the realism of the generated image distribution to the ground truth distribution. Thus, we computed FID between the set of generated images and set of ground-truth images. It is important to note that distortion metrics such as PSNR are not optimized by generative methods. Thus, our diffusion-based approach performs 2-4dB worse than deblurring methods in PSNR (see supplementary) when refocusing, consistent with observations in other generative works~\cite{fleetsuperresolution, superresolution2}. A significant factor contributing to this drop is latent-space compression by the VAE, as our model struggles in the identity case (reconstructing input focus). However, in designing our method, we prioritize perceptual plausibility over exact pixel fidelity, accepting modest losses in reconstruction accuracy to achieve more visually appealing results. 

In Table~\ref{tab:main_benchmark}, we compare our method’s performance on refocusing tasks across different focal positions in the focal stack: the first (position 1), the middle (position 5), and the last (position 9).
Our SVD-based approach achieves superior perceptual quality for larger focal position shifts, such as refocusing from positions 1 or 9. However, at position 5, where many scenes are naturally in focus, and rich in high-frequency details, baseline deblurring networks perform better due to their direct operation in the pixel domain, preserving these details more effectively. By contrast, VAE compression in our method struggles with such fine-grained reconstruction. We observe our method outperforms IP2P and Swintormer, which suggests that image diffusion may be missing priors that video diffusion models have on focal stacks or image diffusion may require a separate contribution with a carefully constructed conditioning setup.

Despite challenges with high-frequency details at small refocusing distances, our results highlight the value of generative methods for refocusing over large refocusing distances, particularly in real-world scenarios where users frequently shift focus between the foreground and background. 

\paragraph{Gamma and noise sensitivity.}
We additionally find our method to be robust under variations in tone mapping (gamma) and noise levels. Specifically, we test two gamma correction settings ($\gamma = 0.5$ and $\gamma = 2$) and additive Gaussian noise with $\sigma = 0.05$ applied to the input images. Quantitative and qualitative results are reported in the supplementary.

\paragraph{User study.} Additionally, we conduct a user study at refocusing distances greater than 4 positions (as this is the regime in which our method performs well). We show users the input image and two refocused versions: NAFNet (the best baseline) and ours. We then ask users to vote for the version they prefer. Order is randomized per trial. We used 20 random scenes and had 20 users. Our method received 88.25\% of votes and NAFNet got 11.75\% of votes. This study shows our method generates user-preferred refocusing results.

\paragraph{Ablation study.} We explore an alternative way of conditioning on the input focal position by replacing our position-dependent scheme with the original motion profile ID from SVD~\cite{svd}. In this ablation, we assign the motion profile ID to the input focal position and replicate the input image latent across all frames as in the original SVD approach. As seen in Table~\ref{tab:ablation_table}, this setup fails to learn to refocus and merely reproduces the input image in every video frame. This ablation highlights the necessity of our position-dependent conditioning with modified CFG for generating valid focal stacks.


  \begin{figure*}[]
\centering
\includegraphics[width=0.91\linewidth]{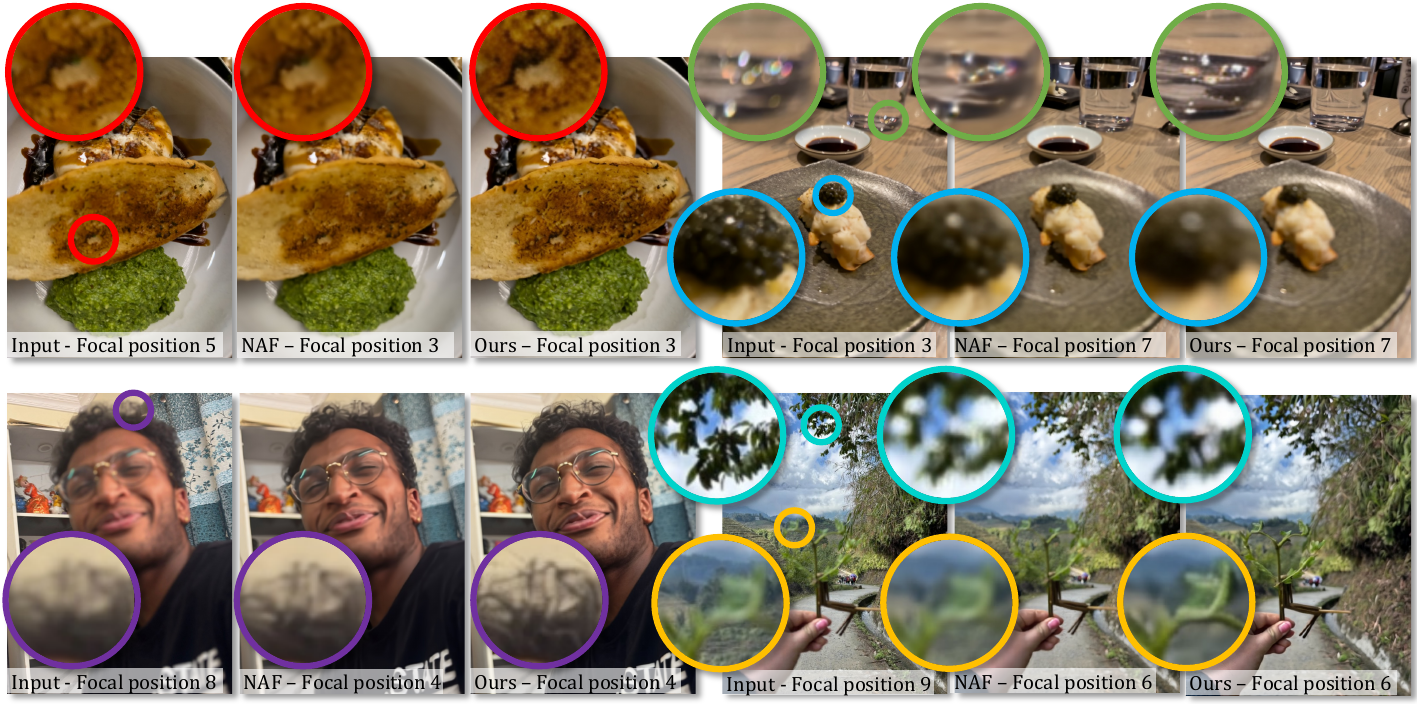}
\caption{Qualitative results of refocusing images provided by everyday iPhone users. We compare against the best baseline, NAFNet~\cite{naf}. Our method has superior refocusing results and handles hard scenes with humans (bottom left) or thin structures (bottom right).}
\label{fig:everyday_results}
\end{figure*}

\subsection{Results on everyday photos}
\label{sec:experiments_everyday_user}

We aim to evaluate our model on scenes collected from casual users, so we gathered a small set of images from everyday iPhone users and tested our method.
When focal position metadata was available, we used the frame closest to the captured focal position as the conditioning signal. 
When metadata was unavailable, we sampled the model by placing the conditioning frame at each focal position and selected the output that produced the most reasonable result through manual inspection. Additionally, we find our model works well with rough estimates of input focal position (near is position 1 and far is position 9), and we discuss further in the supplementary. Refocusing results for these user-provided images are in Figure~\ref{fig:everyday_results}. 

For instance, when refocusing the bread (top left), our method reveals a natural char on the bread's surface. In another example, caustics in the glass (top right) are corrected during refocusing, and bokeh effects from reflected lights in the glass are effectively reconstructed.
Our method also performs well on scenes with faces, preserving fine details in hair (bottom left).
Finally, we demonstrate our model's capability in a challenging scene (bottom right), where the user struggled to focus on a thin plant in their hand; our method successfully focuses the desired subject despite the difficulty.

\begin{figure*}[]
\centering
\includegraphics[width=0.91\linewidth]{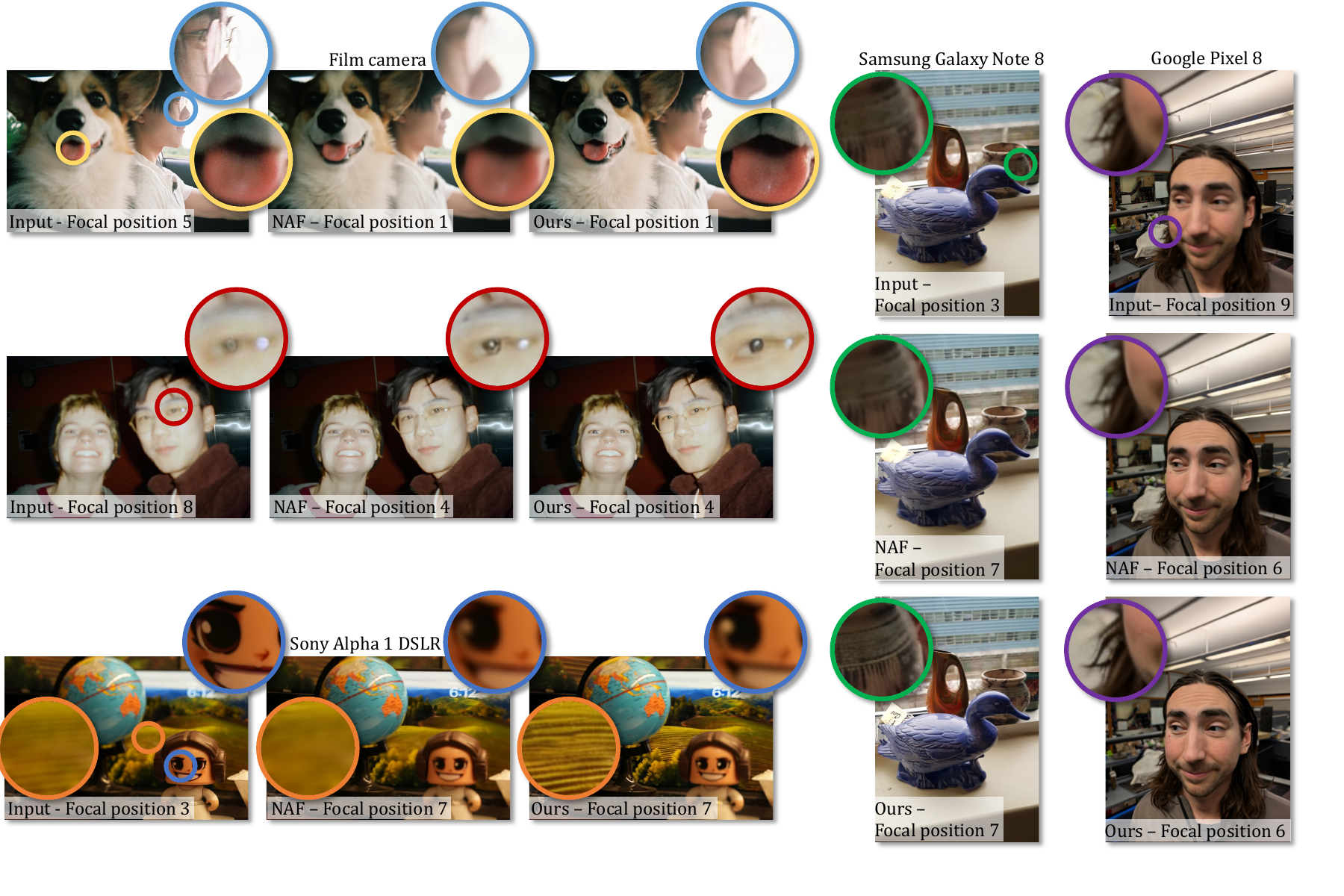}
\caption{
Qualitative results showcasing our method's ability to generalize across different cameras, including film cameras, DSLRs, and smartphones. We compare against the best baseline, NAFNet~\cite{naf}. Our method produces more realistic results than NAFNet across all the cameras.}
\label{fig:other_cameras}
\end{figure*}
\subsection{Generalization to other cameras} 
\label{sec:experiments_other_cameras}
Our method generalizes well to images captured by other smartphone cameras, film cameras, and DSLRs. We demonstrate refocusing results on real-life scenes captured with these devices. Because the focal position metadata for these devices does not align with iPhone focal positions, we apply the same procedure described in Section~\ref{sec:experiments_everyday_user} to refocus their images. Sample results from various cameras are shown in Figure~\ref{fig:other_cameras}. Our method qualitatively outperforms state-of-the-art methods across all examples and reconstructs details in eyes (left col., middle row), hair (right col.), and pottery (middle col.), while preserving the original color profiles of the photos. We visualize additional qualitative comparisons on lightfield datasets in the supplementary.



\subsection{Downstream tasks}
\label{sec:experiments_downstream}
\begin{figure}[]
    \centering
    \includegraphics[width=1\linewidth]{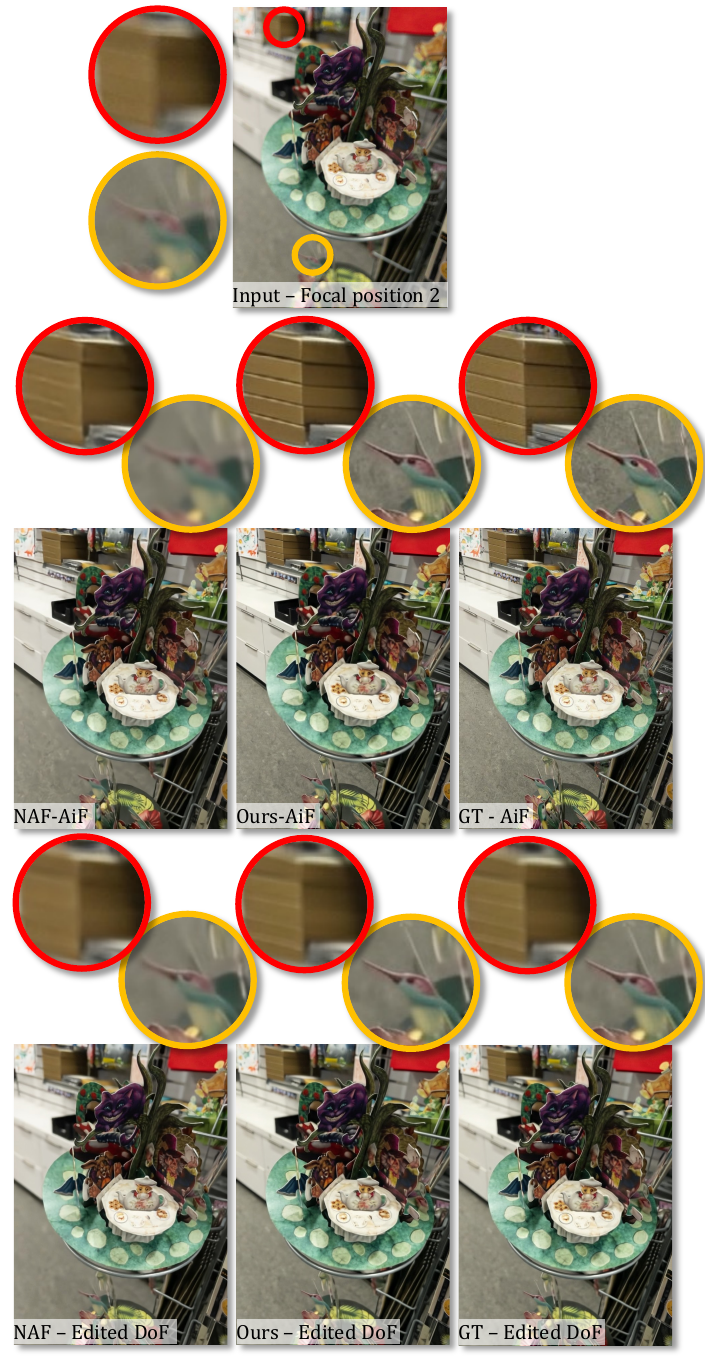}
    \caption{Illustration of generating all-in-focus (AiF) and depth-of-field (DoF) edited images via HeliconFocus applied to our reconstructed focal stacks. For AiF synthesis, we use all frames in the stack; for DoF edits, we use only focal positions 1-5. We compare these AiF/DoF edited outputs from our method against the best-performing baseline, NAFNet~\cite{naf}, as well as the ground-truth (GT) reference.}
    \label{fig:dof_editing}
  \end{figure}

\paragraph{Depth-of-field manipulation.}
Our method’s ability to generate full focal stacks in one shot enables a range of depth-of-field (DoF) editing applications, including creating all-in-focus (AiF) images. We use HeliconFocus to generate AiF images by providing the full focal stack or selectively control the DoF by passing a subset of the focal stack. An example of defocus editing is visualized in Figure~\ref{fig:dof_editing}, where our method produces more perceptually accurate AiF and modified DoF images compared to the baseline. Our approach effectively reconstructs coarse details, such as the spaces between the boxes, and fine details, like the coloring on the bird’s beak.

\begin{figure}[]
\centering
\includegraphics[width=1\linewidth]{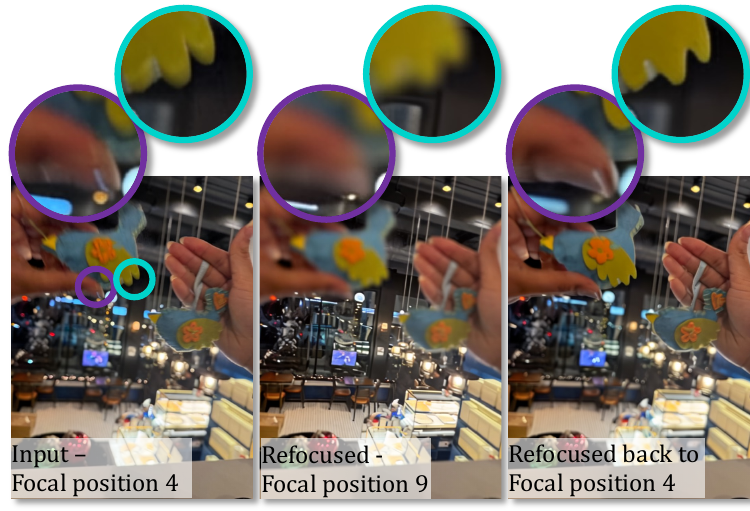}
\caption{We observe that our method can mitigate slight motion blur when refocusing. In the first image, both the hand and bird lie within the depth of field yet appear blurry due to motion. By refocusing to the background (focal position 9), these elements become noticeably sharper. When we subsequently refocus this output back to the original focal position, the hand and bird retain their improved clarity, indicating that the motion blur is substantially reduced.}
\label{fig:motion_blur}
\end{figure}

\paragraph{Motion Deblurring.}
Our method can also handle cases with slight motion blur when performing refocusing. Figure~\ref{fig:motion_blur} shows an example with foreground motion blur from scene motion. When refocusing to bring the background into focus, we observe that the foreground motion blur is reduced. Refocusing the image back to its original focus further reduces the motion blur. Notably, the nail of the person holding the bird and the bird’s wing appear sharp, with minimal motion blur after refocusing twice. Our intuition for this behavior is that diffusion models learn to generate images within the data distribution trained on. Since our dataset contains images without motion blur, our model also learns to generate images without motion blur.

\subsection{Failure cases and limitations} 
\label{sec:experiments_limitations}

\begin{figure}[h!]
    \centering
    \includegraphics[width=1\linewidth]{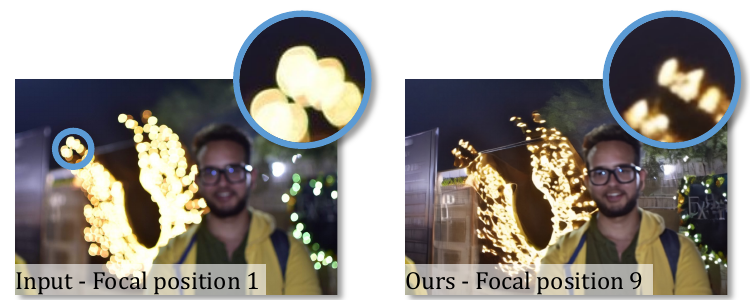}
    \caption{Our method struggles with extreme defocus levels that exceed those present in our training dataset. When attempting to refocus in such cases, it cannot accurately recover fine details—like the illuminated lights—due to the excessively large bokeh. }
    \label{fig:failure_cases}
  \end{figure}

\paragraph{Failure cases.} Our method does not generalize very well to scenes exhibiting extreme defocus or bokeh effects, which are common in images captured by DSLRs with big apertures. Such levels of defocus are not present in our iPhone training dataset, limiting the model's ability to generalize to these scenarios. As shown in Figure~\ref{fig:failure_cases}, our method fails to correct the exaggerated bokeh in DSLR images. Nonetheless, we are confident that by later incorporating datasets from large-aperture cameras into the training set, our method can effectively learn to refocus such images and handle the challenges posed by extreme defocus. Finally, our model is limited to refocusing images, and when applied to video frames, temporal consistency is not enforced, leading to temporal artifacts (see supplementary).

\section{Conclusions} 
We present a novel application of video diffusion models to post-capture refocusing, offering a solution to reconstruct realistic focal stacks from single defocused images.
Our approach modifies the classifier-free guidance mechanism and surpasses existing methods in perceptual quality under larger focal position changes and demonstrates robustness across challenging scenarios. It performs well for refocusing images provided by users from everyday settings and generalizes to cameras beyond the training set, such as film cameras, DSLRs, and other smartphones. Beyond refocusing, our method proves effective for DoF editing and motion blur reduction.

We will also release a dataset of 1,637 real focal stacks captured on a smartphone, featuring RAW-format images with realistic lighting variations and everyday environments, providing a valuable benchmark for refocusing methods.  

\paragraph{Future work.}
We now discuss two future directions for this work. First, our method struggles at reconstructing detail at small focal position changes because we utilize a latent diffusion model. We believe this could be improved by using a pixel diffusion model, a latent model with a better autoencoder, or a hybrid. Second, incorporating data from large-aperture cameras and additional sensors could improve generalization to scenarios with various defocus properties, potentially allowing aperture size or focal length as an additional conditioning factor for learning to refocus.

\begin{acks}
A special thanks to Karanpreet Raja for building the focal stack viewer. Additionally the authors would like to thank Siddhu Tedla, Keshav Pandiri, Trevor Canham, Justin So, Giuliana Mariano, and Yashi Uppalapati for providing images.
\end{acks}

\bibliographystyle{ACM-Reference-Format}
\bibliography{sample-bibliography}

\pagebreak

\clearpage
\phantom{This is a fix for image rendering on the last page.}

\end{document}